\documentclass{article}

\usepackage[preprint]{neurips_2024}

\usepackage[utf8]{inputenc} 
\usepackage[T1]{fontenc}    
\usepackage{hyperref}       
\usepackage{url}            
\usepackage{booktabs}       
\usepackage{amsfonts}       
\usepackage{nicefrac}       
\usepackage{microtype}      
\usepackage{xcolor}         
\usepackage{amsmath}
\usepackage{graphicx}
\usepackage{subcaption}
\usepackage{tcolorbox}
\usepackage{multirow}
\usepackage{xspace}

\title{Large Language Model-guided Document Selection}

\author{%
Xiang Kong\thanks{Equal contribution.} \quad Tom Gunter\footnotemark[1] \quad Ruoming Pang \\
Apple\\
\texttt{\{xiang\_kong,tom\_gunter,r\_pang\}@apple.com}\\
}
\newcommand{\lmlarge}{$\mathrm{LM}_{\text{large}}$\xspace}
\newcommand{\blmlarge}{$\mathbf{LM}_{\text{large}}$\xspace}
\newcommand{\lmsmall}{$\mathrm{LM}_{\text{small}}$\xspace}
\newcommand{\blmsmall}{$\mathbf{LM}_{\text{small}}$\xspace}

\begin{document}

\maketitle

\begin{abstract}
Large Language Model (LLM) pre-training exhausts an ever growing compute budget, yet recent research has demonstrated that careful document selection enables comparable model quality with only a fraction of the FLOPs. Inspired by efforts suggesting that domain-specific training document selection is in fact an interpretable process \citep{gunasekar2023textbooks}, as well as research showing that instruction-finetuned LLMs are adept zero-shot data labelers \citep{Gilardi_2023}, we explore a promising direction for scalable general-domain document selection; employing a prompted LLM as a document grader, we distill quality labels into a classifier model, which is applied at scale to a large, and already heavily-filtered, web-crawl-derived corpus autonomously.  Following the guidance of this classifier, we drop 75\% of the corpus and train LLMs on the remaining data. Results across multiple benchmarks show that: 1. Filtering allows us to quality-match a model trained on the full corpus across diverse benchmarks with at most 70\% of the FLOPs, 2. More capable LLM labelers and classifier models lead to better results that are less sensitive to the labeler’s prompt, 3. In-context learning helps to boost the performance of less-capable labeling models. In all cases we use open-source datasets, models, recipes, and evaluation frameworks, so that results can be reproduced by the community.
\end{abstract}

\section{Introduction}
\label{sec:intro}
Large Language Models (LLMs) (\cite{devlin-etal-2019-bert,radfordimproving,radford2019language,brown2020language,touvron2023llama,touvron2023llama2,achiam2023gpt,team2023gemini}) have demonstrated impressive zero and few-shot generalization across a diverse range of downstream language understanding and generation tasks, after extensive training on large text corpora. Encouraged by the fact that model capabilities seem to continue to improve as more data and compute are put to work, with seemingly no end in sight, scaling law efforts have sought to predict model capability given increased compute and data budgets \cite{neuralScalingLaws,chinchilla}.

In \cite{beatingScalingLaw}, the authors comment that these power scaling laws show disappointingly slow scaling (a 100x increase in training compute improves cross-entropy loss by only 0.5 nats in \cite{chinchilla}, for example), and demonstrate empirically that it may be possible to significantly increase the power-law exponent with careful data selection. Separately, substantial effort typically goes into producing an optimal overall dataset mixture for LLM training \cite{doremi}, yet these mixtures always include "high quality" dataset components, which are typically over-sampled (e.g. StackExchange, Wikipedia, or books for the LLaMA model series \cite{touvron2023llama,touvron2023llama2}), suggesting that these components make significant---relative to their size---contributions to downstream model quality.

By necessity of volume, the majority of any LLM pre-training dataset will be derived from a filtered bulk web crawl, with a popular source being the CommonCrawl (CC) \cite{commoncrawl} corpus. Various open-source filtered subsets of CC exist, e.g. \cite{gao2020pile, together2023redpajama, cerebras2023slimpajama}, which have already been pruned to only include the highest quality documents. Efforts to improve data pruning, and by doing so the scaling law efficiency, should therefore focus on better selecting documents from the bulk web crawl.

In this work, we explore a promising direction for general-domain web-crawl pruning by introducing a scalable approach for LLM-guided document selection. Our framework utilizes two large language models, a large instruction-finetuned model (\lmlarge), and a smaller pre-trained language model (\lmsmall). Given a target corpus, we initially leverage the strong zero-shot capabilities of \lmlarge to assess a number of sampled documents in terms of quality and educational value. Subsequently, \lmsmall is finetuned on the quality labels from \lmlarge. Finally, the finetuned \lmsmall is applied to score the full web-crawl corpus. This distillation step allows us to minimize the total amount of compute required for filtering, as for any given prompt and label-set from \lmlarge we can establish whether a larger distillation model (\lmsmall) is required by simply examining the fine-tuning evaluation metrics.

This paper makes the following contributions:
\begin{itemize}
\item We introduce a scalable and general LLM-powered web-crawl document selection framework that does not rely on high-quality reference corpora. 
\item We apply this pipeline to the filtered RPJ-CC~\cite{together2023redpajama} dataset, as a representative high-quality baseline, and demonstrate significant improvements in downstream LLM quality across multiple parameter and compute budgets.
\item We conduct several ablation studies to study the impact of: 1. the labeling prompt, 2. the capability of the labeling model, 3. the capacity of the distillation model.
\end{itemize}

\begin{figure}[!htbp]
\vspace{2mm}
\centering
\begin{tcolorbox}[width=0.8\textwidth,colback=green!2!white,colframe=gray!50!yellow]
    \begin{minipage}{\textwidth}
{\huge }{ 
[Document]\newline

<I am a document.>\newline

[Instruction] In the above we provide a document snippet. The start and end of the snippet may contain only a partial word, as we sliced at the character level. Is the document snippet educational and engaging for a college student studying a STEM subject or the humanities? Answer with "Yes" or "No" without any additional comments.
} {\huge}
\end{minipage}
\end{tcolorbox}
\caption{The prompt used to guide the LLM labeler to assess the quality and education value for an input document.}
\label{fig:prompt}
\end{figure}

\begin{figure}%
    \centering
    \includegraphics[width=0.95\textwidth]{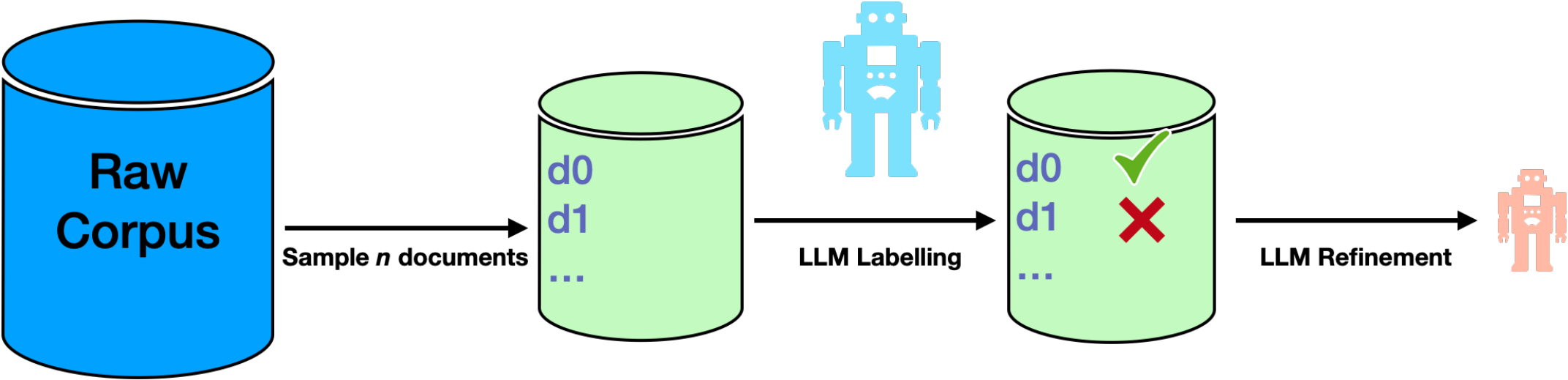}%
    \vspace{3mm}
    \caption{The overall pipeline for our proposed LM-guided data selection pipeline. Given a raw corpus, we first sample $n$ documents and guide an LLM labeler to assess them in terms of the textual quality and educational value. The resulting (doc, label) pairs could be distilled into an LM-based quality classifier, which will label all documents in the raw corpus.}%
    \label{fig:pipeline}%
\end{figure}

\section{Method}
\label{sec:method}
In this section we describe the LLM guided web-crawl document selection (LMDS) pipeline.

\subsection{Proposed Framework}
To apply LMDS, we first need access to the following:
\begin{itemize}
\item \textbf{Target Corpus} ($C$): The source dataset over which document selection is to be applied. $C$ may contain hundreds of billions of documents sourced from the web, (e.g. plaintext extracted CommonCrawl), with text covering a diverse range of topics, languages, etc.
\item \textbf{\blmlarge}: A capable (the more so the better), instruction-finetuned language model.
\item \textbf{\blmsmall}: A second, low inference cost distillation-target LLM, preferably heavily pretrained (instruction finetuning is optional).
\end{itemize}

The pipeline is composed of two distinct stages:
\paragraph{\blmlarge-Labelling} In this stage, we select a random subset of documents from $C$ to serve as a representative sample for the whole corpus. The sampled documents are then fed into \lmlarge for assessment according to the criteria specified in the prompt (Figure~\ref{fig:prompt}), generating high/low-quality document labels. 
We note that the choice of prompt used to guide \lmlarge is important for two reasons: 1. it must be straightforward for the chosen \lmlarge to interpret (very capable proprietary models are often able to follow more subtle instructions versus open-source alternatives), 2. it explicitly defines the selection criteria. Assuming a broad definition of "high quality", we find that a concise prompt guiding \lmlarge to assess documents for general educational value (inspired by the domain-targeted efforts of \cite{gunasekar2023textbooks}) works well. We leave it to future work to explore more targeted (e.g. subject-specific) prompts, but do believe that the flexibility of the prompt (vs. e.g. sourcing relevant high-quality reference corpora to train a classifier) is a major benefit of LMDS.

\paragraph{\blmsmall-Refinement}
To enable labelling at very large scale, a cheap-to-run-inference-on (smaller) pretrained language model \lmsmall is employed. We first distill the labels sourced from \lmlarge on the document assessment task into \lmsmall. Once fine-tuning is complete, \lmsmall can be deployed to evaluate the entire target corpus $C$ using your accelerator-backed map-reduce framework of choice. 

The overall two-stage pipeline is depicted in Figure~\ref{fig:pipeline}.

\paragraph{Rationale for Using \blmsmall as well as \blmlarge}
 Given the relative size of a filtered crawl dataset vs the unfiltered corpus, (typically $\ll 1\%$ of the data remains after selection \cite{together2023redpajama,cerebras2023slimpajama}), we believe that document selection is a difficult recall problem---it's probable, given non-perfect filter recall, that a significant number of high quality documents are dropped by typical filter pipelines.
 For this reason we'd like to minimize the number of stages (asides from LMDS), in the overall filter pipeline, which in turn means that it must be feasible to run LMDS at a hundred-billion document scale.
 At the same time, we want to use the most powerful \lmlarge that we have access to, so that the generated labels adhere to the prompt with high fidelity.
 To balance this trade-off we introduce \lmsmall, and in Table \ref{tab:classifier_size} we show that it's possible to  understand the effect of this distillation step (and as a result tune the size of \lmsmall) by examining the F1 score of the classifier at fine-tuning time, with little to no overhead.

\section{Experiments}
\label{sec:exp}
\subsection{Experiment Setup}
\paragraph{Source dataset:}
To validate the technique, we will use the filtered Common Crawl (CC) split from the RedPajama v1 data mixture (RPJ-CC) \cite{together2023redpajama} as $C$, because this is a reasonably strong baseline over which we hope to improve. We leave it to future work to demonstrate the effectiveness of LMDS applied to a larger scale unfiltered web-crawl (although we expect this to yield better results still).
The RPJ-CC dataset was produced by running five CommonCrawl dumps through the cc\_net pipeline, which de-duplicates (including at a paragraph level) and selects documents using both perplexity filtering and a linear classifier trained to recognize high-quality Wikipedia reference text. It contains 878 billion tokens.

\paragraph{Language Model-guided Document Selection Pipeline:} The data selection pipeline described in Section~\ref{sec:method} is applied to these datasets.

In the \textbf{\blmlarge-Labelling} phase, we use the Llama-2-chat 70B (\cite{touvron2023llama})\footnote{https://huggingface.co/meta-llama/Llama-2-70b-chat-hf} as our \lmlarge, as it is a widely accessible open-source model with good language understanding and reasoning capabilities. We sample two million documents from RPJ-CC and feed the middle 1500 tokens of each document into \lmlarge, (along with the prompt in Figure~\ref{fig:prompt}), for quality assessment\footnote{We set temperature as 0.2 to make the results more deterministic.}. 

During the \textbf{\blmsmall-Refinement} stage, we take \lmsmall (a smaller LLM pretrained on Wikipedia, Stackexchange, and Books components from \citet{cerebras2023slimpajama}) and finetune this model on the (document, quality label) pairs generated by \blmlarge, in the process distilling the signal from \blmlarge. More specifically, a Sigmoid binary classification head is added to \lmsmall, and the representation produced by this head defines the quality score.

At this point \lmsmall is a distilled representation of the labeling function defined by the prompt and \blmlarge, and is then used to quality score $C$.

\paragraph{Language model training on selected documents:} 

To evaluate the efficacy of our data selection pipeline, we train language models of varying capacities (1B and 7B) on a high quality subset of $C$ according to the scores from our \lmsmall. For all documents in the target corpus, \lmsmall assigns a quality score in $(0, 1)$, then we choose a cutoff and keep only documents scoring above this threshold for training. We drop around 75\% of the data from $C$, which is in-line with the fraction of documents assigned to the high-quality bucket by \blmlarge at label generation time. 

We compare models trained with the same compute budget (and using sequence packing \cite{gpt2paper}), i.e. all models see exactly the same number of tokens and are trained with the same hyperparameters. For details about the architecture and training processes, please refer to  Appendix~\ref{sec:appendix_arch}.

\paragraph{Language model evaluation:} We assess the language models across a comprehensive set of benchmarks. For all results reported in our study, higher numbers are better than lower numbers.

\begin{itemize} \item \textbf{Core English tasks (CoreEN):} Models are evaluated across various benchmarks focusing on on common sense reasoning, reading comprehension and question answering, including ARC Easy and Challenge (0-shot) (\cite{clark2018arc-e-c}), HellaSwag (0-shot) (\cite{zellers2019hellaswag}), WinoGrande (0-shot) (\cite{sakaguchi2019winogrande}), PIQA (0-shot) (\cite{bisk2019piqa}), SciQ (0-shot) (\cite{sap2019socialiqa}), LAMBADA (0-shot) (\cite{paperno2016lambada}), TriviaQA (1-shot) (\cite{joshi2017triviaqa}) and WebQS (1-shot) (\cite{Berant2013webqs}). We use the EleutherAI evaluation-harness \cite{eleuther-eval-harness} to run these evaluations. In our experiments, we report both the average performance on the seven 0-shot tasks (denoted as CoreEN (0S)) and the average performance across all nine tasks (denoted as CoreEN (All)).

\item \textbf{MMLU:} Performance is detailed for the 5-shot setup on the MMLU benchmark~(\cite{mmlu}),  a widely-adopted benchmark consisting of exam-style questions from 57 tasks.
\end{itemize}

\begin{figure}[htbp]
    \centering
    \begin{subfigure}[b]{0.33\textwidth}
        \centering
        \includegraphics[width=\textwidth]{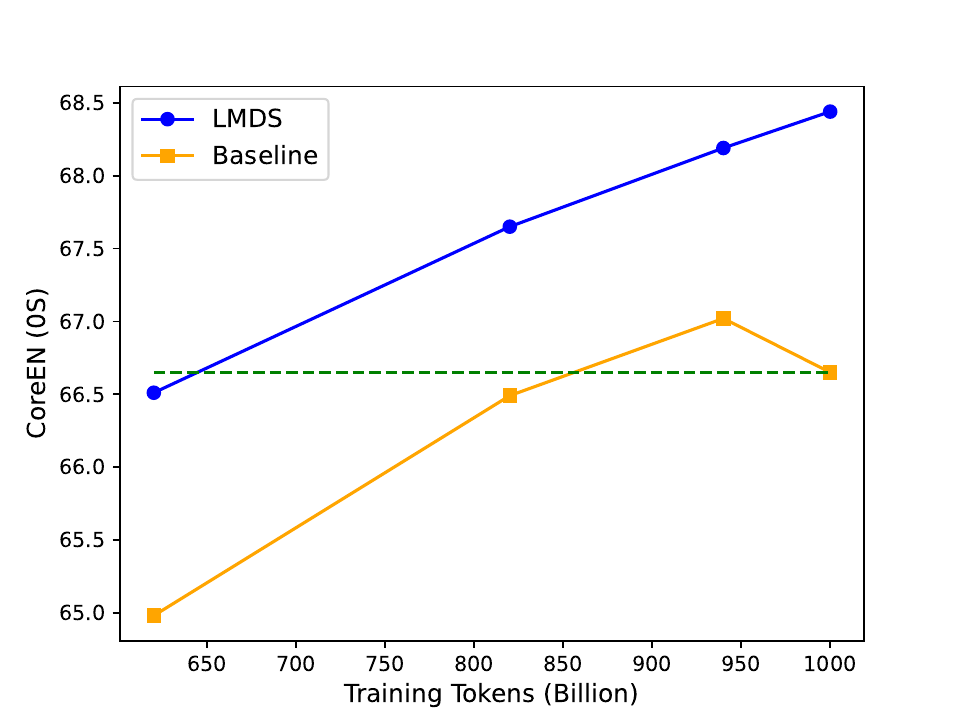}
       \caption{7B CoreEN (0S)}
        \label{fig:subfig_q}
    \end{subfigure}
    \hfill
    \begin{subfigure}[b]{0.32\textwidth}
        \centering
        \includegraphics[width=\textwidth]{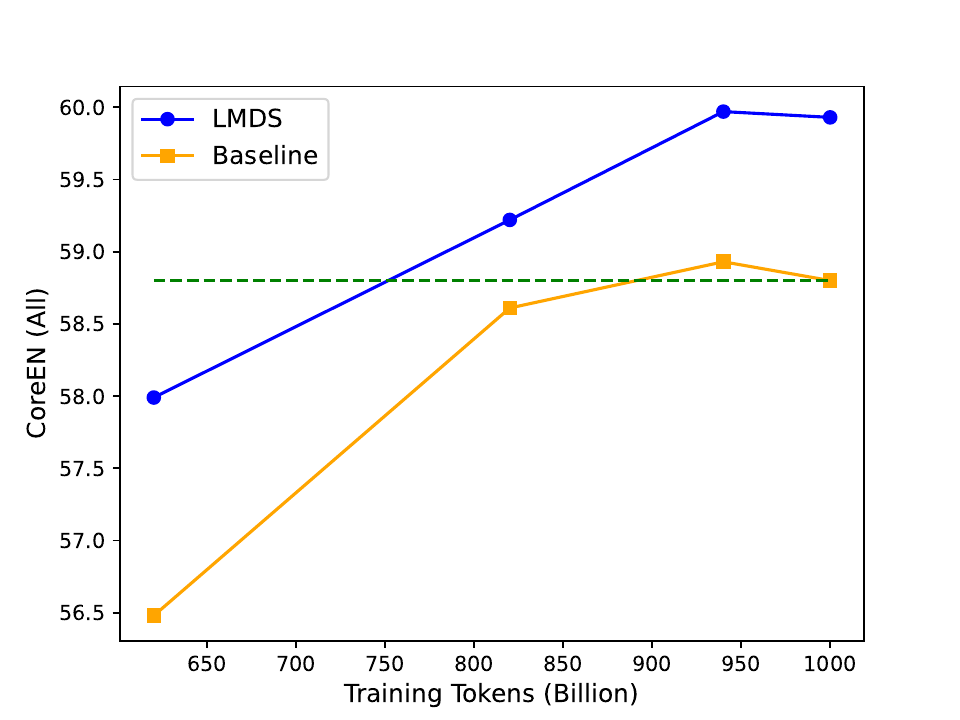}
        \caption{7B CoreEN (All)}
        \label{fig:subfig_b}
    \end{subfigure}
    \hfill
    \begin{subfigure}[b]{0.33\textwidth}
        \centering
        \includegraphics[width=\textwidth]{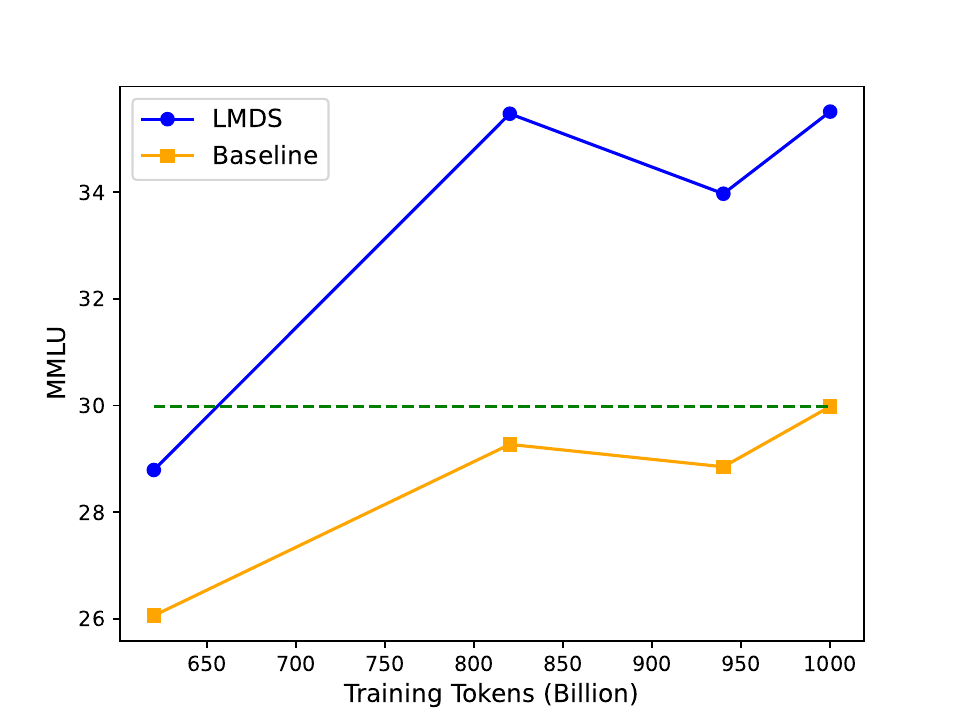}
        \caption{7B MMLU}
        \label{fig:subfig_c}
    \end{subfigure}
    \caption{Learning curves on the downstream task for 7B model pretraining on the raw data versus LMDS-based filtered data.}
    \label{fig:data_efficiency}
\end{figure}
\subsection{Experiment Results}
\label{sec:exp_results}
\begin{table}
  \caption{Comparisons of models trained on datasets w/ (LMDS) and w/o data selection. We do not report MMLU results for the 1B model, because at this compute budget they are near random.}
  \label{tab:main_results}
  \vspace{0.5cm}
  \centering
  \begin{tabular}{c | l | c  c c c}
    \toprule
     model & dataset & CoreEN (0S) & CoreEN (All) & MMLU \\
     \midrule
    1B & Baseline &  58.10 &  48.63 & N/A \\
    1B & LMDS &  \textbf{61.14} & \textbf{51.41} & N/A  \\
    \midrule
    7B & Baseline & 66.66 & 58.72 & 29.98 \\
    7B & LMDS &  \textbf{68.44} & \textbf{59.93} & \textbf{35.51} \\
    \bottomrule
  \end{tabular}
\end{table}

\paragraph{Model quality with a Fixed Training Budget:} In Table~\ref{tab:main_results}, we evaluate models of different scales across multiple benchmarks. Models trained on datasets curated using our proposed data selection pipeline consistently achieve higher performance than their counterparts across all benchmarks. For example, at 7B scale, the model trained on filtered Common Crawl achieves \textbf{+5.53} on MMLU score compared to the one trained on entire RPJ-CC dataset, demonstrating the effectiveness of our method to select documents with high educational value.  The breakdown performance on CoreEN is listed in Appendix~\ref{sec:appendix_break}.

\paragraph{Data efficiency:} Given the same training budget (FLOPs), training on the LMDS-filtered dataset achieves higher downstream model performance.
We further investigates the data efficiency of our approach. As depicted in Figure~\ref{fig:data_efficiency}, we show the performance of 7B models trained on w/ and w/o our LMDS pipeline against the number of training tokens. We find that filtering allows us to quality-match a model trained on the full corpus across diverse benchmarks with at most 70\% of the FLOPs.

\begin{figure}[htbp]
    \centering
    \includegraphics[width=0.7\textwidth]{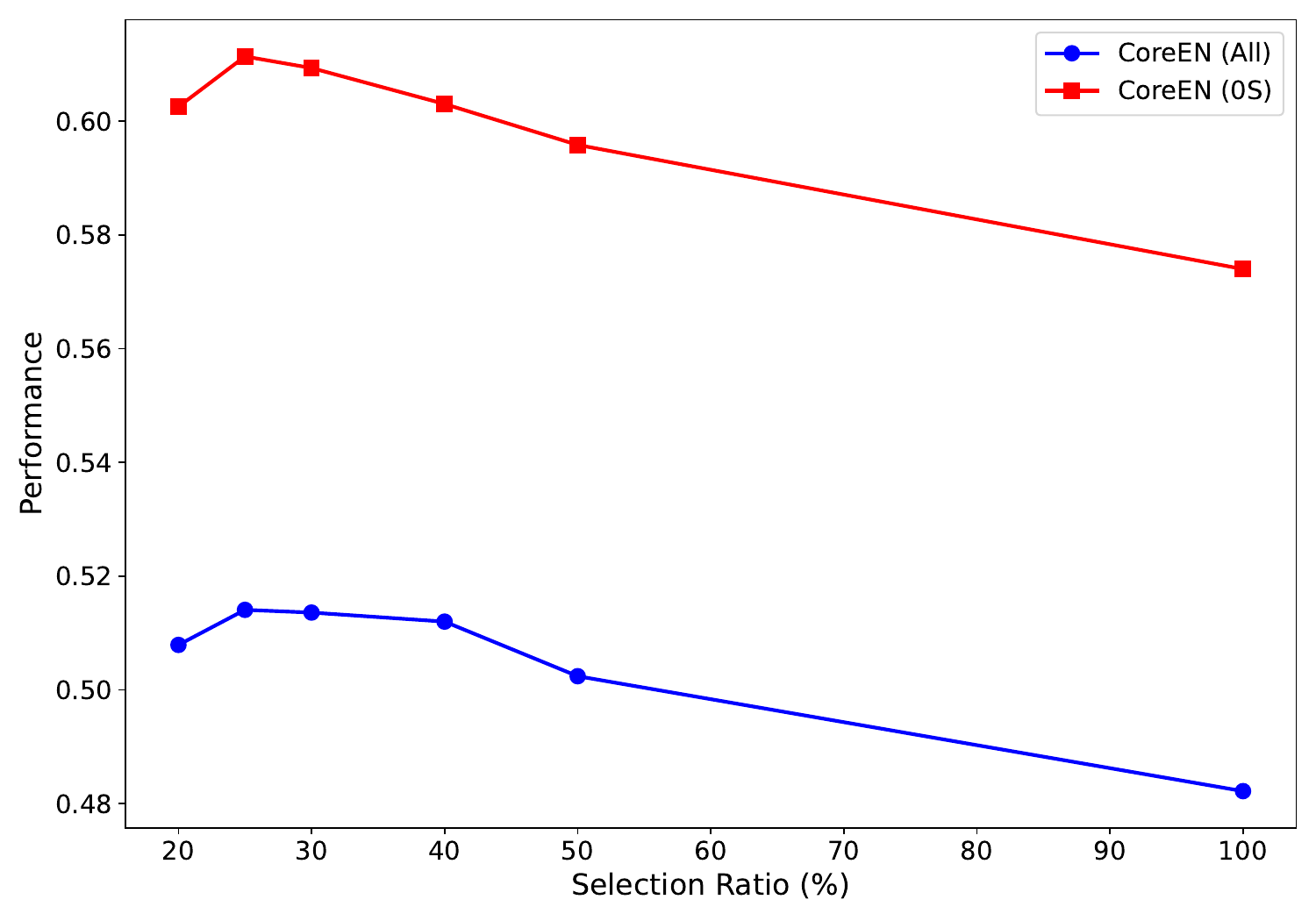}%
    \caption{Downstream task accuracy of 1B models trained on LMDS-based filtered datasets with different selection ratios.}%
    \label{fig:selection_ratio}%
\end{figure}

\paragraph{Downstream Task Accuracy v.s. Selection Ratio:} Using the finetuned \lmsmall to label the target corpus, each document in the target corpus receives a score ranging from zero to one. One natural question is how to set the cutoff threshold to select documents for model training. In this study, we explore cutoff thresholds to achieve distinct selection ratios ($\{20\%, 25\%, 30\%, 40\%, 50\%, 100\%\})$ and train 1B models on these resulted datasets to understand its impact on the downstream task performance.
The results on CoreEN (0S) and CoreEN (All) are shown in Figure~\ref{fig:selection_ratio}. Firstly, all selection ratios $\{20\%, 25\%, 30\%, 40\%, 50\%\}$ show meaningful improvements over the baseline dataset (selection ratio=100\%). Secondly, as the data selection ratio is reduced from 55\% to 25\%, we observe incremental gains in performance, underscoring the importance of data quality to downstream task outcomes, whilst an overly aggressive pruning (reduction from 25\% to 20\%) begins to reduce performance again. Note that the the optimal dataset selection ratio may vary for different corpora, however the raw label-split declared by \lmlarge is a good rule of thumb when selecting an appropriate threshold.

\begin{table}
  \caption{Comparing distillation classifier models across various sizes. Reported F1 score is computed on the validation set. It's clear that 1. a classifier as small as 85m parameters provides a meaningful improvement over the baseline, 2. larger classifiers are still meaningfully better.}
  \vspace{0.5cm}
  \label{tab:classifier_size}
  \centering
  \begin{tabular}{l | c c  c c c}
    \toprule
     Size & F1-score & CoreEN (0S) & CoreEN (All) \\
     \midrule
    85M & 0.78 & 60.24 & 50.72\\
     302M& 0.81 & 60.88 & 51.10\\
    1B & \textbf{0.84} & \textbf{61.14} & \textbf{51.41}\\
    \bottomrule
  \end{tabular}
\end{table}
\paragraph{Downstream Task Accuracy v.s. Size of \blmsmall:} In this study, we examine the impact of different sizes of \lmsmall on downstream task accuracy. Specifically, we experiment with three sizes of \lmsmall: 85M, 302M, and 1B parameters. Our first step involves comparing the F1-scores these models achieve on a held-out set when fine-tuned on the (document, quality label) pairs annotated by \lmlarge. From the results presented in Table~\ref{tab:classifier_size}, it is evident that larger models yield higher F1-scores. This trend suggests that models with greater capacity are more effective at learning from the annotations produced by our large language model, \lmlarge.
To further investigate this phenomenon, we leverage the three fine-tuned \lmsmall models to label the target corpus. We then train 1B LMs on these datasets. The result is shown in Table~\ref{tab:classifier_size}, we observe that the LM trained on the dataset labeled by the 1B \lmsmall outperforms those trained on datasets labeled by the 302M and 85M \lmsmall models, showing that larger capacity \lmsmall can learn better during the distillation stage. We conclude that models as small as 85m parameters provide a meaningful improvment over the baseline, and that larger models do better still---so suggest that the choice of \lmsmall capacity is guided by the total inference budget available.
 
\begin{table}
  \caption{Comparing the different scale $\mathrm{LM}_{\text{large}}$ on downstream task accuracy.}
  \vspace{0.5cm}
  \label{table:llm_large_size}
  \centering
  \begin{tabular}{l | c c  c}
    \toprule
     $\mathrm{LM}_{\text{large}}$ &  CoreEN (0S) & CoreEN (All) \\
     \midrule
     Baseline & 58.10 & 48.63 \\
    Llama-2-chat 7B & 57.44 & 48.13 \\
     Llama-2-chat 13B & 59.50 & 50.39\\
    Llama-2-chat 70B &   \textbf{61.14} & \textbf{51.41}\\
    \bottomrule
  \end{tabular}
\end{table}
\begin{table}
  \centering
  \caption{The robustness of $\mathrm{LM}_{\text{large}}$ with various sizes to prompt instructions. We train models based on three versions of instructions for $\mathrm{LM}_{\text{large}}$ to label sample documents.}
  \vspace{5mm}
  \label{tab:robust}
  \begin{tabular}{l|c|c|c}
    \toprule
    $\mathrm{LM}_{\text{large}}$ & Prompt Version & CoreEN (All) & Avg. (Std) \\
    \midrule
    \multirow{3}{*}{Llama-2-chat 13B} & V1 & 50.05 & \multirow{3}{*}{$50.27_{\pm 0.57}$} \\
                                      & V2 & 50.86 &  \\
                                      & V3 & 49.90 &  \\
    
    \midrule
    \multirow{3}{*}{Llama-2-chat 70B} & V1 & 51.41 & \multirow{3}{*}{$51.24_{\pm 0.24}$} \\
                                      & V2 & 51.34 &  \\
                                      & V3 & 50.96 &  \\
    \bottomrule
  \end{tabular}
\end{table}
\paragraph{Impact of \blmlarge:} Another critical component of our data selection pipeline involves crafting effective prompts that guide large language models \lmlarge accurately tag documents with respect to their quality and educational value. To achieve this, \lmlarge needs to have strong instruction following, textual understand and reasoning capabilities. Therefore, in this study, we first try three \lmlarge with different scales, i.e., Llama-2-chat 7B, 13B and 70B~\cite{touvron2023llama}. We first employ these three models to label the same sample documents from the target corpus and finetune \lmsmall on them. We train 1B LMs on data labeled by these \lmsmall. From the Table~\ref{table:llm_large_size}, we find that the larger \lmlarge is, the better accuracy we can obtain. For the Llama-2-chat 7B, it only achieve similar accuracy compared to the baseline and after a closer look, we find that approximately 98\% of the sample documents are labeled as \textit{yes}, indicating that the Llama-2-chat 7B model lacks the ability to analyze intricate and nuanced textual content effectively.

Furthermore, we also try to understand the robustness of Llama-2-chat 13 and 70B models with respect to instructions. Ideally, \lmlarge should provide a consistent quality assessment given semantically similar instructions.  Therefore, we try three prompts (more details in Appendix~\ref{sec:appendix_prompts}), and compute the agreement of downstream evaluation results across these prompts. As shown in Table~\ref{tab:robust}, the Llama-2-chat 70B model exhibits higher agreement compared to the Llama-2-chat 13B, indicating that larger capability \lmlarge is more reliable and less influenced by variations in prompts.

\paragraph{In-Context Learning for Llama-2-chat 7B:} Due to its limited instruction-following capabilities, Llama-2-chat 7B struggles to distinguish high-quality documents from low-quality ones, as compared to its larger siblings. However, large language models exhibit in-context learning (ICL) abilities (\cite{brown2020language}), where they can learn from a few examples provided in the context without fine-tuning and by doing so overcome instruction-following shortcomings. To leverage this, we provide 5 examples from the strongest \lmlarge Llama-2-chat 70B in this study as a demonstration context for Llama-2-chat 7B. We then trained 1B models based on the ICL-enhanced Llama-2-chat 7B results. As shown in Table~\ref{table:icl}, incorporating demonstrations from a stronger \lmlarge yields better downstream accuracy, showing ICL helps 
the Llama-2-chat 7B to more effectively identify high-quality documents. However, there still remains a significant gap compared to \lmlarge, indicating that raw model capability (especially instruction following) is critical for accurate document labeling.

\begin{table}
  \caption{Comparing the Llama-2-chat 7B w/ and w/o in-context learning on downstream task accuracy.}
  \vspace{0.5cm}
  \label{table:icl}
  \centering
  \begin{tabular}{l | c c  c}
    \toprule
     $\mathrm{LM}_{\text{large}}$ &  CoreEN (0S) & CoreEN (All) \\
     \midrule
    Llama-2-chat 7B & 57.44 & 48.13 \\
     ~~~~~+ ICL (5-shot) & 58.55 & 48.80 \\
     Llama-2-chat 70B & 61.14 & 51.41 \\
    \bottomrule
  \end{tabular}
\end{table}

\section{Related Work}
\label{sec:related}

The goal of data filtering is to determine the optimal subset of training data to train on, where optimal is typically measured via trained model performance on a diverse suite of benchmarks. It is well known that better data selection leads to improved downstream performance and/or a reduced compute budget to achieve the same performance for large-scale model training efforts, e.g. LLMs \cite{beatingScalingLaw}.

We believe that the most closely related work to our own is \cite{sachdeva2024train}, (developed concurrently), where the authors employ a T5 \cite{raffel2020exploring} model to produce quality labels before training new T5 models on the filtered data via a process they term "Ask-LLM". 
Compared to Ask-LLM, we focus on decoder-only LLMs, demonstrating that the technique works for much larger models and compute budgets (7B parameters trained for 1T tokens), as well as tougher benchmarks (MMLU), and also works when starting with a stronger baseline dataset in RPJ-CC (versus C4\cite{raffel2020exploring}). Furthermore, we ablate the impact of the prompt as well as the choice of labeler and distillation model, showing that more capable labelers are both more effective and less sensitive to the choice of prompt.

In what follows we describe more broadly approaches for data selection by quality, with a focus on efforts applied to language model training.

\paragraph{Effect of data selection on pre-training stage:} 

Training large language models typically involves utilizing extensive text datasets sourced from massive and diverse origins such as CommonCrawl and GitHub. However, it is widely recognized that pre-training data is very noisy, and often includes boilerplate text, templates, error messages, and other forms of repetitive or not-informative content that does not contribute meaningfully to model quality as measured by downstream benchmarks. This poor quality data can lead models to learn and perpetuate irrelevant or redundant information, ultimately impacting their performance and generalization capabilities across various tasks (\cite{raffel2020exploring,touvron2023llama,touvron2023llama2}). For instance, during the creation of the Pile dataset (\cite{gao2020pile}), authors find that the most common 13-grams are character repetitions, such as strings of dashes, with more than 10 million instances. Removing such undesirable text is crucial but must be done efficiently due to the large number of documents involved. A common approach to filtering data is to employ simple yet computationally efficient heuristics. The goal of these heuristic approaches is to constrain the training distribution along certain dimensions (e.g., sentence length, repetitiveness), with the assumption that the evaluation distribution will exhibit similar characteristics. The heuristics used in past works are extensive but generally fall into one of the following categories: item count, repetition count, ratio, or per-document statistical measures.

In addition to heuristic data selection methods, many works focus on developing more advanced model-based techniques to identify and select "high quality" text. \citet{brown2020language,du2022glam,touvron2023llama,together2023redpajama} try to select data in a noisy dataset based on the similarity to a pre-defined high quality reference corpus, e.g., Wikipedia, to improve the model quality. \cite{gao2020pile} train GPT-2 (\cite{achiam2023gpt}) and GPT-3 (\cite{brown2020language}) models from scratch to compare the "Pile" dataset with CommonCrawl to demonstrate the effectiveness of data filtering. \citet{raffel2020exploring} train 1B parameter models to show the positive impact of the de-duplication and quality filters. \citet{hernandez2022scaling} train multiple language models with various epoch counts but fixed compute budgets to understand the effect of repeated data. \citet{muennighoff2024scaling} explore the effect of dataset repetition on scaling laws, finding that repeating the highest quality data for up to four epochs outperforms fresh, less-heavily filtered data seen only once. In \citet{tirumala2024d4}, the authors conduct a large-scale data-efficient pre-training evaluation, showing that a 6.7B OPT model (\cite{zhang2022opt} can converge up to 20\% faster on data curated by a technique based on pre-trained model embeddings on top of de-duplicated data. Instead of hard-selection, \citet{xie2023data} propose a data selection alogrithm with importance resampling to estimates importance weights and select data according to these weights during the LLM training. One line of research focuses on the the data mixture between multiple components. \citet{longpre2023pretrainer} discuss the effect of several factors including toxicity, age and domain distribution on the downstream performance. \citet{xie2024doremi} propose to seek data mixture ratios through training a small proxy model using Group DRO, which could be used to train large scales models. Another series of studies focuses on training models using a selection of "textbook quality" data to demonstrate the importance of data quality (\cite{javaheripi2023phi,gunasekar2023textbooks,li2023textbooks,eldan2023tinystories}). In, \citet{maini2024rephrasing}, the authors use off-the-shelf instruction-finetuned language models to paraphrase documents on the web to improve the data quality without selection, and as previously mentioned, \citet{sachdeva2024train} employs T5 to choose the high quality data with density sampling to preserve the coverage and diversity of the training data at the same time. \cite{zhang2024autonomous} employs a language model to help identify domain specific (Math) documents to boost math-related benchmark performance.

\paragraph{Effect of data selection on post-training stage:}
In contrast to the pre-training, the post-training stage aims to align LLMs with a downstream user's objective to make the model outputs more controllable and helpful for users (\citet{ouyang2022training}). Many works aim to improve the data quality through focusing on the  difficulty, complexity and diversity as components of the utility. \citet{zhou2024lima} show that only 1,000 carefully curated data with high quality and diversity can achieve superior instruction following capabilities. \citet{li2023quantity} design a difficulty metric from instruction-following perspective to let model itself to select the training samples. \citet{kung2023active} propose \textit{activate instruction tuning} to identify informative tasks to continuously improve its cross-task generalization ability. Although human guidance may be more reliable, recently, the capability of models to precisely identify undesirable data points is also effective. For example, \citet{chen2023alpagasus} leverage ChatGPT \cite{achiam2023gpt} to assess training examples and select a subset of Alpaca dataset. \citet{gunasekar2023textbooks} build an LM-based classifier to filter textbook quality data from the target corpus and this classifier is trained on data generated from LLMs. Some off-the-shelf LLMs such as GPT-4 (\cite{achiam2023gpt} are used to filter and rank responses to provide high-quality feedback (\cite{cui2023ultrafeedback,zhu2023starling}).

\section{Conclusion} \label{sec:conclusion}
This paper introduces a scalable and general large language model-guided document pruning pipeline for the autonomous selection of high-quality training data for large language models (LLMs) from a web-scale corpus. Two language models with different scales, \lmlarge and \lmsmall achieve a good balance of selection precision/recall and inference efficiency/scalability. We demonstrate the effectiveness of this approach by applying it to the already-filtered RPJ-CC dataset, at multiple model size and training FLOP budgets. Results showed that models trained on the filtered dataset achieve significantly better quality across multiple benchmarks, in most cases doing so with a reduced compute budget. Several ablation studies highlight the importance of key components in our framework. In future work we hope to demonstrate the effectiveness of this pipeline applied at full scale to a (near) unfiltered web-corpus, and expect that domain targeted \lmlarge prompts will also allow for the targeted selection of domain-specific data for domains which do not have an existing high-quality and high-coverage reference corpus.

\newpage
\newpage
\bibliographystyle{plainnat}
\bibliography{main}
\newpage
\appendix
\section{Modeling Training Details}
\label{sec:appendix_arch}
$\mathbf{LM}_{\text{small}}$ pretraining: Before finetuning $\mathbf{LM}_{\text{small}}$ on (document, label) pair labelled by $\mathbf{LM}_{\text{large}}$, we first pretrain it on Wikipedia, Stackexchange, and Books components from \citet{cerebras2023slimpajama} to build a good initialization for finetuning. 

The architecture of all language models including $\mathbf{LM}_{\text{small}}$ and $\mathbf{LM}_{\text{large}}$is listed in the Table~\ref{tab:all_architectures}. All models are pretrained with the AdamW optimizer~(\cite{loshchilov2019decoupled}), employing parameters $\beta_1=0.9$, $\beta_2=0.95$, decoupled $wd=1e-4$ and $\epsilon = 10^{-8}$. We implement a cosine learning rate schedule, starting with 2000 warmup steps and decaying to 10\% of peak learning rate and a gradient clipping at 1.0. A SentencePiece tokenizer~(\cite{kudo2018sentencepiece}), employing a vocabulary of 32k and built on the Byte-Pair Encoding algorithm (\cite{sennrich2016bpe}), is used on C4. All the model are trained on TPU-v4 and v5 chips. All $\mathbf{LM}_{\text{small}}$  are trained with 400B tokens. 1B and 7B main models are trained with 300B and 1T tokens respectively. We use the $\mu$Param (simple) (\cite{wortsman2023small}) where the peak learning rate is set to 1e-2.

\begin{table}
  \caption{All model specifications used in this paper.}
  \vspace{2mm}
  \label{tab:all_architectures}
  \centering
  \begin{tabular}{l | c c c c}
    \toprule
    model &  num layers & hidden dim & num heads & batch size (\# tokens)\\
    \midrule
    85M & 12 & 768 & 12  & 1M \\
    302M & 24 & 1024 & 16 & 1M \\
    1B & 24 & 2048 & 32  & 1M \\
    7B & 32 & 4096 & 32 & 4M \\
    \bottomrule
  \end{tabular}
\end{table}
\section{Breakdown performance on CoreEN tasks}
\label{sec:appendix_break}
We list performance of different models on CoreEN tasks in Table~\ref{tab:breakdown_results_1b}.
\begin{table}[htbp]
  \caption{Breakdown results of 1B models on CoreEN in Table~\ref{tab:main_results}}
  \label{tab:breakdown_results_1b}
  \vspace{0.5cm}
  \centering
  \begin{tabular}{l |  c c}
    \toprule
      & Baseline & LMDS \\
     \midrule
     arc\_c & 26.37 & 35.07\\
     arc\_e  & 62.21 & 70.54\\
     hellaswag  &43.75 & 43.16 \\
     lambada  & 56.44 & 58.88\\
     piqa  & 71.22 & 69.59 \\
     sciq  & 86.80 & 90.20 \\
     winogrande  & 59.91 & 60.54 \\
     triviaqa  & 17.68 & 20.99\\
     webqs  &13.29 & 13.68\\
    \bottomrule
  \end{tabular}
\end{table}

\begin{table}[htbp]
  \caption{Breakdown results of 7B models on CoreEN in Table~\ref{tab:main_results}}
  \label{tab:breakdown_results_7b}
  \vspace{0.5cm}
  \centering
  \begin{tabular}{l |  c c}
    \toprule
      & Baseline & LMDS \\
     \midrule
     arc\_c & 40.02 & 44.45\\
     arc\_e  & 72.06 & 77.10\\
     hellaswag  &50.88 & 52.04 \\
     lambada  & 70.25 & 69.75\\
     piqa  & 73.78 & 74.65 \\
     sciq  & 92.50 & 94.30\\
     winogrande  & 66.06 & 66.77 \\
     triviaqa  & 40.79 & 39.13\\
     webqs  &21.90& 21.16\\
    \bottomrule
  \end{tabular}
\end{table}

\section{Instructions exploration for $\mathbf{LM}_{\text{large}}$}
We list the prompts we used to test the robustness of different $\mathbf{LM}_{\text{large}}$. They are semantically similar to each other and a strong language model should be less sensitive to these prompts. 
\label{sec:appendix_prompts}
\begin{table}[htbp]

\centering

\begin{tabular}{|p{0.5cm}|p{13cm}|}

\hline

V1 & \textbf{[Instruction]} In the above we provide a document snippet. The start and end of the snippet may contain only a partial word, as we sliced at the character level. Is the document snippet educational and engaging for a college student studying a STEM subject or the humanities? Answer with "Yes" or "No" without any additional comments. \\ \hline

V2 & \textbf{[Instruction]} In the above we provide a document snippet. The start and end of the snippet may contain only a partial word, as we sliced at the character level. Does the document look like it would be helpful for a STEM or Humanities student who is struggling with their course? Answer with "Yes" or "No" without any additional comments. \\ \hline

V3 & \textbf{[Instruction]} In the above we provide a document snippet. The start and end of the snippet may contain only a partial word, as we sliced at the character level. Does the document look like it would be educational and helpful for a STEM or Humanities student to help understanding material from their course? Answer with "Yes" or "No" without any additional comments. \\ \hline

\end{tabular}

\caption{Prompts used to test the robustness of different $\mathbf{LM}_{\text{large}}$.}

\label{table:instructions}

\end{table}
\clearpage
\end{document}